\documentclass[11pt]{article}

\usepackage[final]{acl}

\usepackage{times}
\usepackage{latexsym}
\usepackage[T1]{fontenc}
\usepackage[utf8]{inputenc}
\usepackage{microtype}
\usepackage{graphicx}
\usepackage{amsmath,amssymb,amsthm}

\usepackage{booktabs}
\usepackage{multirow}
\usepackage{float}
\usepackage{enumitem}
\usepackage{xcolor}
\usepackage{xspace}
\usepackage{tikz}
\usetikzlibrary{positioning,arrows.meta,fit,backgrounds,calc}
\usepackage{tcolorbox}
\tcbuselibrary{breakable}

\definecolor{maincolor}{HTML}{1f77b4}

\newtcolorbox{takeawaybox}{
  colback=maincolor!6,
  colframe=maincolor!75!black,
  width=\columnwidth,
  boxrule=0.6pt,
  leftrule=2pt,
  arc=0.8mm,
  auto outer arc,
  left=4pt,
  right=4pt,
  top=3pt,
  bottom=3pt,
  before skip=5pt,
  after skip=5pt,
  fontupper=\small,
  breakable,
}

\definecolor{cellred}{HTML}{E41A1C}
\definecolor{cellblue}{HTML}{377EB8}
\definecolor{cellgreen}{HTML}{4DAF4A}
\title{AWM: Answerable Working Memory for Long-Document VQA Agents}

\author{
  \textbf{Dongzhuoran Zhou\textsuperscript{1,5,*}},
  \textbf{Yuqicheng Zhu\textsuperscript{2,5}},
  \textbf{Yule Liu\textsuperscript{3,*}},
  \textbf{Zhen Yang\textsuperscript{4}},
\\
  \textbf{Rui Lu\textsuperscript{4}},
  \textbf{Yuxiao Dong\textsuperscript{4}},
  \textbf{Jie Tang\textsuperscript{4}},
  \textbf{Evgeny Kharlamov\textsuperscript{1,5}}
\\
\\
  \textsuperscript{1}University of Oslo,
  \textsuperscript{2}University of Stuttgart
\\
  \textsuperscript{3}Hong Kong University of Science and Technology (Guangzhou)
\\
  \textsuperscript{4}Tsinghua University,
  \textsuperscript{5}Bosch Center for AI
}

\begin{document}
\maketitle
\begingroup
\renewcommand{\thefootnote}{\fnsymbol{footnote}}
\footnotetext[1]{\raggedright Corresponding authors.\par
Code is available at\,
\href{https://github.com/DongzhuoranZhou/AWM}{github.com/DongzhuoranZhou/AWM}\par}
\endgroup

\begin{abstract}
Long-document visual question answering increasingly relies on VLM agents that retrieve candidate pages, inspect page images, write findings to working memory, and synthesize answers.
Working memory should carry answer-supporting evidence across page inspections for later grounded answering, yet existing evaluation mainly checks final-answer correctness and evidence-page access.
This creates a memory-quality blind spot: an agent may reach the right page and answer correctly while leaving behind memory too generic or incomplete to support answering once page context is removed.
We introduce \emph{memory-only answerability}, a diagnostic that asks whether a reader can answer from the question and terminal working memory alone.
Building on this diagnostic, \emph{Answerable Working Memory} (AWM) treats terminal working memory as an answerable evidence artifact, and AWM-GRPO incorporates this signal into the GRPO reward while preserving final-answer priority.
Under GRPO, this reward assigns higher advantages to answer-correct trajectories whose terminal working memory remains answerable.
On \textsc{MMLongBench-Doc}, even when gold evidence pages are provided, 42.5\% of correct answers still cannot be answered from terminal working memory alone.
AWM-GRPO improves final-answer accuracy over the RAG baseline by 8.1 and 11.9 points on \textsc{MMLongBench-Doc} and \textsc{LongDocURL} and reduces the memory-missing-correct rate by 2.7 points over answer-only GRPO.
\end{abstract}

\section{Introduction}
\label{sec:intro}

Answering questions over long documents resembles goal-directed information seeking: a reader searches for relevant evidence, keeps intermediate findings available, and integrates them into an answer.
Recent long-document VQA systems instantiate a similar loop with VLM agents: they retrieve candidate pages, inspect page images, update working memory with findings, and synthesize a final answer~\citep{colpali,visrag,m3docrag,docvstar,mmdocr1,scopevlm}.
Working memory should carry answer-supporting evidence across page inspections for later grounded answering.
Yet current evaluation checks final-answer correctness and evidence-page access, but not whether terminal working memory remains answer-supporting.
An agent may therefore answer correctly from page context while leaving behind memory too generic to support the answer on its own.

We introduce \emph{memory-only answerability}: a reader receives only the question and the agent-written terminal working memory, without page images or trajectory context, and must answer from that memory alone.
This exposes a memory-quality gap: even when gold evidence pages are supplied, 42.5\% of correctly answered \textsc{MMLongBench-Doc} examples fail memory-only answerability.
A controlled memory comparison further shows that improving terminal working memory improves answering, motivating optimization of the memory artifact itself.

Motivated by this observation, we propose AWM-GRPO.
For each trajectory, the agent outputs a final answer and terminal working memory.
A frozen reader answers from the question and terminal working memory alone, and a judge scores both the final answer and the memory-only answer.
Under GRPO, this reward ranks trajectories by both final-answer correctness and memory answerability, creating a preference for trajectories that answer correctly with answerable terminal working memory.

We evaluate a controlled training pipeline in which the three AWM-Agent variants share the same Qwen3-VL-4B policy, RAG Top-3 retrieval harness, tools, memory schema, prompts, and optimizer; only post-training and reward differ.
We compare direct input, RAG Top-3, SFT, answer-only GRPO, and AWM-GRPO.
On \textsc{MMLongBench-Doc}/\textsc{LongDocURL}, AWM-GRPO exceeds RAG Top-3 by 8.1/11.9 points, SFT by 4.7/4.4 points, and answer-only GRPO by 2.3/2.7 points.
It also reduces the memory-missing-correct rate by 2.7 points over answer-only GRPO on \textsc{MMLongBench-Doc} (Sec.~\ref{subsec:analysis}).

This paper makes three contributions.
(1) We identify a memory-quality gap in long-document VQA agents: final-answer correctness and evidence-page access do not determine whether terminal working memory preserves answer-supporting evidence.
(2) We propose AWM-GRPO, which uses memory-only answerability as a GRPO reward signal and ranks trajectories by both final-answer correctness and memory answerability, creating a preference for answer-correct trajectories with answerable terminal working memory.
(3) We evaluate AWM-GRPO against direct-input, RAG, SFT, and answer-only GRPO baselines, showing improvements in both final-answer accuracy and memory-quality diagnostics.

\section{Agent Setup and Memory-Quality Motivation}
\label{sec:diagnostic}

This section introduces the agent-memory framework and the memory-only answerability metrics used throughout the paper.
It then presents two diagnostics: current agent-written memory is often insufficient even when page access is controlled, and improving terminal working memory improves answering.

\begin{figure*}[t]
  \centering
  \includegraphics[width=\textwidth]{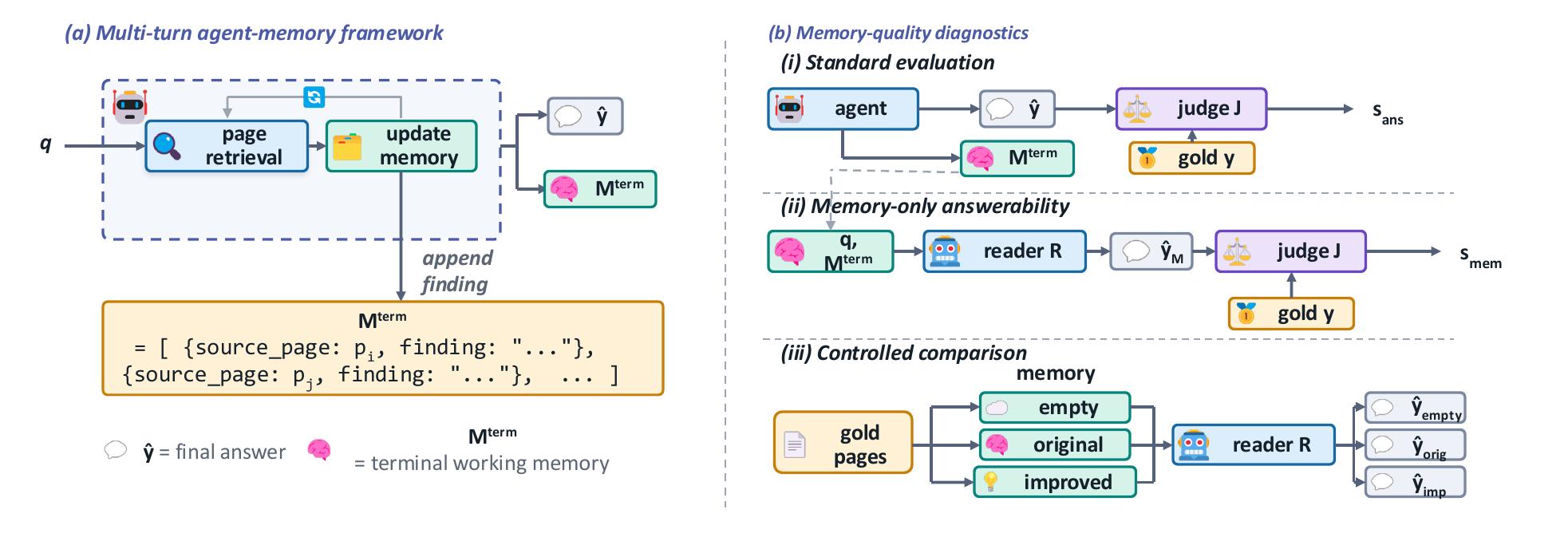}
  \caption{Combined agent-memory framework and memory-quality diagnostics. (a)~The agent retrieves page images, writes source-linked findings to working memory, and outputs both a final answer and terminal working memory. (b)~Standard evaluation scores the final answer, memory-only evaluation asks whether the question can be answered from terminal working memory alone, and the controlled comparison varies terminal working memory while holding evidence-page access fixed.}
  \label{fig:agent-diagnostic}
\end{figure*}

\paragraph{Agent-memory framework.}
\label{sec:setup}
A long-document VQA instance is a tuple $(D, q, y)$, where $D = \{p_1, \ldots, p_N\}$ is a multi-page document, $q$ is a question, and $y$ is the ground-truth answer; when available, $E \subseteq D$ denotes gold evidence pages.
The policy $\pi$ uses evidence-page retrieval and memory update.
A retrieval call returns candidate page images directly in the observation; after inspecting these images, the agent must append a source-linked finding to working memory $M_t$.
The interaction proceeds until the agent outputs both a final answer $\hat{y}$ and terminal working memory $M^{\mathrm{term}}$, or reaches a step budget (Figure~\ref{fig:agent-diagnostic}a).

As illustrated in Figure~\ref{fig:agent-diagnostic}a, memory entries are source-linked findings whose quality depends on whether they preserve question-relevant evidence.
A generic description such as ``the chart shows trends'' is less useful than a question-specific finding such as ``the 2020 self-sufficiency rate is 4.7\%.''

\paragraph{Memory-only answerability.}
To measure whether terminal working memory $M^{\mathrm{term}}$ alone supports answering, we evaluate \emph{memory-only answerability}: a frozen reader $R$ receives only the question $q$ and $M^{\mathrm{term}}$, without the trajectory or page images, and produces a memory-only answer $R(q, M^{\mathrm{term}})$.
The official judge $J$ evaluates both the agent's final answer and the reader's memory-only answer against the gold answer $y$.
For the reported four-cell analysis, we map any positive score from $J$ to $1$ and zero to $0$.
This gives two binary scores per instance: final-answer correctness $s_{\mathrm{ans}} = \mathbf{1}[J(\hat{y}, y)>0]$ and memory-only answerability $s_{\mathrm{mem}} = \mathbf{1}[J(R(q, M^{\mathrm{term}}), y)>0]$.
Implementation details for $R$ and $J$ appear in Appendix~\ref{app:judge}.

The pair $(s_{\mathrm{ans}}, s_{\mathrm{mem}})$ partitions instances into four cases (Table~\ref{tab:cells2x2}). The focal case is \textbf{memory-missing correct} (MMC): $s_{\mathrm{ans}} = 1$ and $s_{\mathrm{mem}} = 0$, i.e., the agent answered correctly but its terminal working memory alone does not support answering.

\begin{table}[t]
\centering
\small
\resizebox{\columnwidth}{!}{%
\begin{tabular}{l l l}
\toprule
Final answer & Memory answerable & Case name \\
\midrule
correct & yes & memory-supported correct \\
correct & no  & \textbf{memory-missing correct} \\
wrong   & yes & answering error \\
wrong   & no  & unresolved error \\
\bottomrule
\end{tabular}}
\caption{Outcome cases induced by final-answer correctness and memory-only answerability.}
\label{tab:cells2x2}
\end{table}

We summarize each run with $P_{\mathrm{mmc}}=\frac{\sum_i \mathbf{1}[s_{\mathrm{ans}}^{(i)}{=}1 \,\wedge\, s_{\mathrm{mem}}^{(i)}{=}0]}{\sum_i \mathbf{1}[s_{\mathrm{ans}}^{(i)}{=}1]}$, the fraction of final-answer-correct instances whose terminal working memory is not answerable.
Conditioning on $s_{\mathrm{ans}}{=}1$ removes general answer failure as a confound and asks whether memory remains insufficient even after the agent answered correctly.

\paragraph{Current memory quality.}
We measure $P_{\mathrm{mmc}}$ on a fixed 500-example answerable subset of \textsc{MMLongBench-Doc} (excluding the benchmark's unanswerable questions).
The \emph{full multi-turn} setting runs the agent end-to-end with its own retrieval and memory-update process.
The \emph{evidence-page-given} (EP-given) setting replaces retrieval with gold evidence pages, so page access is controlled and retrieval is no longer a confound.
Table~\ref{tab:memquality} reports the results.

\begin{table}[t]
\centering
\small
\setlength{\tabcolsep}{3pt}
\resizebox{\columnwidth}{!}{%
\begin{tabular}{l r r r r r}
\toprule
Setting & $N$ & \shortstack[r]{Final-ans.\\acc.\ (\%)} & \shortstack[r]{Memory-only\\acc.\ (\%)} & \shortstack[r]{MMC /\\correct} & $P_{\mathrm{mmc}}$ (\%) \\
\midrule
Full multi-turn & 500 & 33.4 & 27.4 & 43 / 172 & 25.0 \\
EP-given        & 500 & 36.2 & 30.2 & 77 / 181 & 42.5 \\
\bottomrule
\end{tabular}}
\caption{Memory-quality diagnostic for the Base AWM-Agent (Qwen3-VL-4B, no post-training) on a fixed 500-example answerable \textsc{MMLongBench-Doc} subset. Unanswerable questions are excluded. Accuracy follows the official mean per-question score; for MMC/correct and $P_{\mathrm{mmc}}$, we binarize the official score by treating a positive score as correct.}
\label{tab:memquality}
\end{table}

In the full multi-turn setting, one in four correct trajectories ($P_{\mathrm{mmc}} = 25.0\%$) leaves behind terminal working memory that cannot support answering on its own.
The EP-given setting controls for page access by replacing retrieval with gold evidence pages; $P_{\mathrm{mmc}}$ rises to $42.5\%$, showing that page access and final-answer correctness do not imply answerable terminal working memory.

\paragraph{Controlled memory comparison.}
\label{sec:oracle-intervention}
We next ask whether higher-quality memory improves answering.
We construct a controlled setting in which retrieval returns gold evidence pages and compare three terminal-memory variants: (1) \emph{empty memory}; (2) \emph{original agent memory}; and (3) \emph{improved memory}, generated from the same gold pages by a stronger model.
To isolate memory quality, the comparison keeps gold-page access and the evaluation protocol fixed while substituting different terminal working memories.
The answer generator, memory-only reader, and judge are shared across conditions, so only terminal-memory content changes (Figure~\ref{fig:agent-diagnostic}b).
Appendix~\ref{app:oracle} gives the full setup, confidence intervals, outcome-cell breakdown, and artifact audit.

\begin{table}[t]
\centering
\small
\resizebox{\columnwidth}{!}{%
\begin{tabular}{l c c}
\toprule
Memory condition & \shortstack[c]{Final-ans.\\binary rate} & \shortstack[c]{Memory-only\\binary rate} \\
\midrule
Empty memory                 & 2.8  & 2.6  \\
Original agent memory (4B)   & 36.2 & 30.2 \\
Improved memory (GPT-4o)     & 44.4 & 44.8 \\
\bottomrule
\end{tabular}}
\caption{Controlled memory comparison on a fixed 500-example \textsc{MMLongBench-Doc} EP-given answerable subset. The reader and scorer are fixed; only the terminal-memory condition changes. Improved memory is produced by GPT-4o.}
\label{tab:oracle-intervention}
\end{table}

Table~\ref{tab:oracle-intervention} isolates the role of terminal memory.
With empty memory, both final-answer and memory-only binary rates are near zero.
Reusing the 4B agent's original memory yields final-answer and memory-only binary rates of 36.2 and 30.2, while GPT-4o-improved memory raises them to 44.4 and 44.8.
Improving terminal memory alone therefore improves both readouts with the answer generator, reader, and judge held fixed.

\begin{takeawaybox}
\textbf{Takeaway.} Page access and final-answer correctness do not guarantee answerable terminal working memory. Memory-only answerability exposes this gap, and the controlled comparison shows that improving terminal working memory alone can improve answering.
\end{takeawaybox}

\section{Optimizing Answerable Working Memory with GRPO}
\label{sec:method}

\begin{figure*}[t]
\centering
\includegraphics[width=\textwidth]{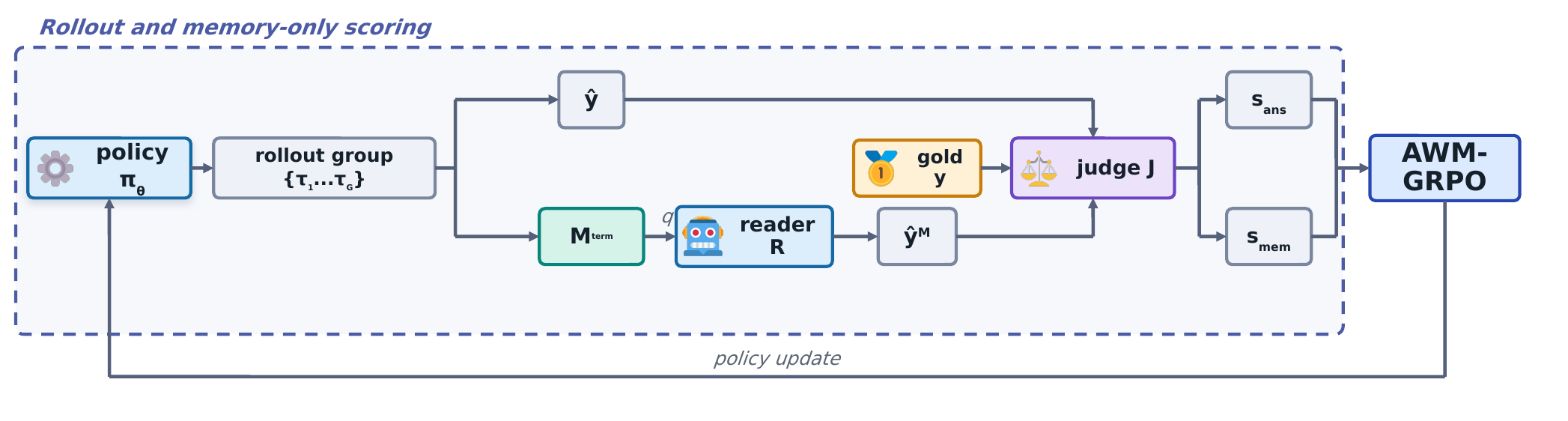}
\caption{Overview of AWM-GRPO. For each sampled trajectory, the agent produces both a final answer and terminal working memory. A frozen reader answers from the terminal working memory alone, and a frozen scorer evaluates the final answer and the memory-only answer. The two scores define the AWM reward, which is normalized within each GRPO group to update the policy.}
\label{fig:method}
\end{figure*}

Sec.~\ref{sec:diagnostic} identifies the failure mode that AWM-GRPO targets: trajectories can answer correctly while leaving terminal working memory unanswerable.
Since GRPO updates the policy through group-relative advantages, the reward should make this memory-quality distinction visible in the advantage ranking.
We first specify the desired advantage behavior, then instantiate a four-cell reward that realizes it, and finally use synthetic outcome mixtures to show how the induced preferences change with group composition.

\subsection{Desired advantage behavior under GRPO}
\label{subsec:advantage-analysis}

For each prompt, GRPO~\citep{shao2024deepseekmath} samples $G$ trajectories and normalizes each trajectory's reward within the group:
\[
A_i = \frac{r(\tau_i) - \mu_g}{\sigma_g + \epsilon},
\]
where $\mu_g$ and $\sigma_g$ are the group mean and standard deviation and $\epsilon = 10^{-6}$ stabilizes zero-variance groups. Because GRPO updates through these normalized advantages, what matters is not the raw reward scale but the \emph{ordering} it induces: for any two outcome cells $z, z'$,
\[
A_z - A_{z'} = \frac{r_z - r_{z'}}{\sigma_g + \epsilon}.
\]
The reward ordering therefore determines the advantage ordering. We design the AWM reward to induce three pairwise preferences:

\paragraph{Memory refinement.}
Among answer-correct trajectories, those whose terminal working memory remains answerable should receive higher advantage than those whose memory is not answerable.

\paragraph{Final-answer gate.}
Final-answer correctness remains primary: an answer-correct but memory-imperfect trajectory should still outrank an answer-wrong trajectory, even if the latter leaves answerable memory.

\paragraph{Useful-memory failure.}
An answer-wrong trajectory with answerable memory is not success, but it is less severe than complete failure, because useful evidence was preserved.

GRPO gives positive advantage only to cells whose reward exceeds the current group mean. The same cell can therefore change role as group composition changes: answer-correct but memory-imperfect trajectories may be tolerated when most samples fail, but become disfavored once memory-supported-correct trajectories are common. Formally, $A_z > 0$ iff $r_z > \mu_g$.

\subsection{AWM reward instantiation}
\label{subsec:reward}

To realize these preferences with a minimal scalar reward, we reuse the two binary variables from Sec.~\ref{sec:diagnostic}: final-answer correctness $s_{\mathrm{ans}}$ and memory-only answerability $s_{\mathrm{mem}}$.
During online training, a frozen local Qwen3-14B model generates the memory-only answer and scores both answers for the reward.
The official judge $J$ from Sec.~\ref{sec:diagnostic} is used for reported diagnostics.
The AWM reward is defined over the resulting four outcome cells:
\begin{gather*}
r_{11}=\beta,\quad r_{10}=0,\quad r_{01}=\gamma,\quad r_{00}=\omega,\\
\beta>0>\gamma>\omega.
\end{gather*}
We write $R_{\mathrm{AWM}}(s_{\mathrm{ans}},s_{\mathrm{mem}})=r_{s_{\mathrm{ans}}s_{\mathrm{mem}}}$; the answer-only baseline uses $R_{\mathrm{ans}}(s_{\mathrm{ans}})=s_{\mathrm{ans}}$.
Here $\beta$ is the reward for full success (memory-supported correct), $\gamma$ is the reward for answer-wrong but memory-answerable trajectories, and $\omega$ is the reward for complete failure. The anchor $r_{10}=0$ fixes answer-correct but memory-imperfect trajectories. The resulting normalized margins are:
\begin{center}
\scriptsize
\setlength{\tabcolsep}{2pt}
\begin{tabular}{lll}
\toprule
Preference & Margin & Ranking \\
\midrule
Memory refinement     & $A_{11}{-}A_{10}=\beta/(\sigma_g{+}\epsilon)$              & $(1,1)\succ(1,0)$ \\
Final-answer gate      & $A_{10}{-}A_{01}=-\gamma/(\sigma_g{+}\epsilon)$            & $(1,0)\succ(0,1)$ \\
Useful-memory failure  & $A_{01}{-}A_{00}=(\gamma{-}\omega)/(\sigma_g{+}\epsilon)$  & $(0,1)\succ(0,0)$ \\
\bottomrule
\end{tabular}
\end{center}
The memory-refinement margin is exactly what answer-only reward cannot express: it scores $(1,1)$ and $(1,0)$ identically, collapsing $A^{\mathrm{ans}}_{11}=A^{\mathrm{ans}}_{10}$.

Each cell's advantage sign depends on the group mean ($A_z > 0$ iff $r_z > \mu_g$):
\begin{center}
\footnotesize
\setlength{\tabcolsep}{4pt}
\begin{tabular}{ccl}
\toprule
Cell & Positive iff & Interpretation \\
\midrule
$(1,1)$ & $\mu_g<\beta$   & full success favored \\
$(1,0)$ & $\mu_g<0$       & flips at $\mu_g{=}0$ \\
$(0,1)$ & $\mu_g<\gamma$  & flips at $\mu_g{=}\gamma$ \\
$(0,0)$ & never            & zero only if $\mu_g{=}\omega$ \\
\bottomrule
\end{tabular}
\end{center}

The main experiments instantiate $\beta=2$, $\gamma=-0.1$, and $\omega=-1$, yielding $(r_{11}, r_{10}, r_{01}, r_{00})=(2,\ 0,\ -0.1,\ -1)$. Under this default, the wrong-answer but memory-answerable cell ($r_{01}{=}\gamma{=}{-}0.1$) is slightly below the answer-correct / memory-imperfect cell ($r_{10}{=}0$), so it becomes negative first as the group mean rises. When the group mean exceeds $r_{10}$, the answer-correct / memory-imperfect cell also becomes negative and positive advantage concentrates on full success. This gives the reward an answer-first, memory-refinement-later tendency without implying a deterministic curriculum guarantee.

\subsection{Advantage distributions under synthetic outcome mixtures}

\begin{figure}[t]
\centering
\includegraphics[width=\columnwidth]{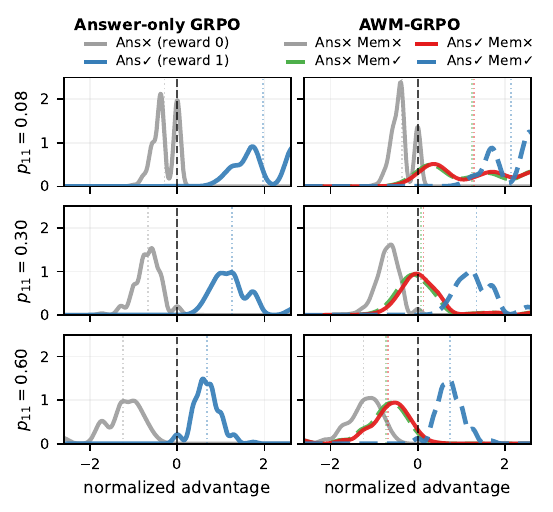}
\caption{Simulated GRPO advantage distributions. Each row samples synthetic groups of size $G{=}8$ from a four-cell outcome mixture; $p_{11}$ is the probability of memory-supported correct $(1,1)$. Left: answer-only reward collapses the four cells into answer-wrong and answer-correct regions. Right: AWM assigns a distinct raw reward to each of the four cells before group normalization.}
\label{fig:advantage-dist}
\end{figure}

Figure~\ref{fig:advantage-dist} uses a controlled Monte Carlo simulation to show how the answer-only and AWM rewards translate into GRPO advantages.
For each row, we fix a categorical distribution over the four outcome cells of Table~\ref{tab:cells2x2} and draw 15{,}000 independent GRPO groups, each containing 8 trajectories.
The answer-only simulation samples from the corresponding final-answer marginal, while the AWM simulation samples the four cells directly.
Rows increase $p_{11}$, the probability of memory-supported correct $(1,1)$, so later rows contain more memory-supported correct trajectories.
The simulation uses the default reward instantiation $\beta=2$, $\gamma=-0.1$, $\omega=-1$.
It isolates reward-induced group-relative preferences; curves are smoothed for visualization only.
Implementation details appear in Appendix~\ref{app:design}.

\noindent\textbf{Three effects are visible:}
\begin{itemize}[topsep=2pt,itemsep=1pt,leftmargin=1.5em]
\item \textbf{Answer-only hides memory quality}: in the left column, trajectories are grouped only by final-answer correctness, so {\color{cellblue}Ans$\checkmark$ Mem$\checkmark$} and {\color{cellred}Ans$\checkmark$ Mem$\times$} are learned together as a single answer-correct region; memory answerability is invisible.
\item \textbf{AWM distinguishes the answer-correct cells}: in the right column, {\color{cellred}Ans$\checkmark$ Mem$\times$} and {\color{cellblue}Ans$\checkmark$ Mem$\checkmark$} receive distinct raw rewards and can therefore receive different normalized advantages. This is the memory-refinement preference missing from answer-only reward.
\item \textbf{The preference changes with group composition}: as $p_{11}$ increases, the red {\color{cellred}Ans$\checkmark$ Mem$\times$} curve moves below zero while the blue {\color{cellblue}Ans$\checkmark$ Mem$\checkmark$} curve remains positive. The nearby green {\color{cellgreen}Ans$\times$ Mem$\checkmark$} curve is expected because $r_{01}{=}{-}0.1$ is close to $r_{10}{=}0$; the key comparison is between the two answer-correct cells.
\end{itemize}

\paragraph{What the reward does not claim.}
The reward uses memory-only answerability as a training signal, not as a claim of full source-grounding verification.
It requires no gold terminal working memory: the frozen reader $R$ answers from the agent-written $M^{\mathrm{term}}$, and the frozen training scorer evaluates both the agent's final answer and the memory-only answer against the gold answer.
The reward does not score memory style, length, or formatting, and memory-only answerability does not override final-answer correctness; $r_{10}>r_{01}$ preserves final-answer priority.
Appendix~\ref{app:design} discusses reward variants, simulation details, and why we adopt this minimal form.

\section{Experiments}
\label{sec:experiments}

We evaluate whether AWM-GRPO improves final-answer accuracy and terminal-working-memory quality over direct-input, RAG, SFT, and answer-only GRPO baselines.

\subsection{Experimental Setup}
\label{subsec:experimental-setup}

\paragraph{Benchmarks.}
We evaluate on \textsc{LongDocURL}~\citep{longdocurl} ($N{=}2325$) and \textsc{MMLongBench-Doc}~\citep{mmlongbenchdoc} ($N{=}1082$), under each benchmark's official evaluation protocol (Appendix~\ref{app:datasets}).

\paragraph{Metrics.}
We report final-answer accuracy, memory-only accuracy, and $P_{\mathrm{mmc}}$ from Sec.~\ref{sec:diagnostic}.
Both final and memory-only answers are evaluated by the official judge $J$ defined in Sec.~\ref{sec:diagnostic}.
The judge uses GPT-4o to extract a concise candidate answer and then applies each benchmark's deterministic rules to compute generalized accuracy.

\paragraph{Agent and retrieval harness.}
The policy is Qwen3-VL-4B-Instruct served via SGLang.
The agent runs the retrieve-inspect-update memory loop of Sec.~\ref{sec:setup} with an append-only mandatory memory-update policy and a maximum of $15$ agent steps per question.
Page retrieval uses Jina v4 page-image embeddings indexed by Qdrant, returning the top three candidate pages per query (RAG Top-3).

\paragraph{Post-training data.}
SFT and GRPO use post-training prompts from \textsc{Doc-750K}~\citep{docopilot}, disjoint from the evaluation sets.
A $2000$-question rendered subset yields $994$ SFT trajectories via difficulty-balanced bucket sampling, with kimi-k2.5 as the teacher; the remaining $1006$ questions form the RL-unseen pool, from which $226$ mixed-difficulty groups are retained for GRPO training.
Online GRPO uses a frozen, locally hosted Qwen3-14B model to generate the memory-only answer and compute both reward scores; only the 4B policy is updated.

\paragraph{Variants.}
We evaluate two no-memory baselines and three AWM-Agent variants.
The \emph{direct-input} baseline receives all document pages in one pass, without retrieval, tools, or terminal working memory.
The second, \emph{VLM + RAG Top-3}, retrieves the top three pages and answers directly without writing terminal working memory.
The three AWM-Agent variants share the same model, tools, memory schema, retrieval harness, prompts, and optimizer; only the post-training step or reward differs: \emph{(i) AWM-Agent (SFT)}: trained on the $994$-trajectory \textsc{Doc-750K} subset.
\emph{(ii) AWM-Agent + Answer-GRPO}: the SFT checkpoint with GRPO under the answer-only reward $R_{\mathrm{ans}}$.
\emph{(iii) AWM-Agent + AWM-GRPO}: the SFT checkpoint with GRPO under the AWM reward $R_{\mathrm{AWM}}$ at default values $(2,\ 0,\ -0.1,\ -1)$ (Sec.~\ref{subsec:reward}).

\subsection{Main Results}
\label{subsec:main-results}

Table~\ref{tab:main-results} reports final-answer accuracy for external baselines and our variants.
The external rows use different backbones and published protocols, so they provide context rather than controlled comparisons.
Our method comparisons use the five Qwen3-VL-4B rows, which share the backbone and official scoring protocol.

\begin{table*}[t]
\centering
\small
\setlength{\tabcolsep}{4pt}
\begin{tabular}{@{}l l c l c c@{}}
\toprule
Method & Backbone & Param & Paradigm & \shortstack[c]{\textsc{MMLongBench-Doc}} & \shortstack[c]{\textsc{LongDocURL}} \\
\midrule
\multicolumn{6}{l}{\emph{Open-source / RAG}} \\
Qwen2.5-VL~\citep{docvstar}              & Qwen2.5-VL & 7B & VLM           & 28.0 & 32.9 \\
SV-RAG~\citep{svrag}                     & InternVL2  & 4B & RAG           & 23.0 & --   \\
M3DocRAG~\citep{m3docrag}                & Qwen2-VL   & 7B & RAG           & 21.0 & 35.1 \\
VisRAG~\citep{visrag}                    & MiniCPM-V  & 8B & RAG           & 18.8 & 41.9 \\
MoLoRAG~\citep{molorag}                  & Qwen2.5-VL & 7B & RAG           & 41.0 & 51.9 \\
URaG-7B~\citep{urag}                     & Qwen2.5-VL & 7B & RAG           & 33.8 & 52.2 \\
\midrule
\multicolumn{6}{l}{\emph{Agentic / RL}} \\
VRAG-RL~\citep{vragrl}                   & Qwen2.5-VL & 7B & Agent + RL    & 26.6 & 44.9 \\
Doc-V$^{\star}$ (SFT)~\citep{docvstar}   & Qwen2.5-VL & 7B & Agent + SFT   & 39.8 & 53.0 \\
Doc-V$^{\star}$ (GRPO)~\citep{docvstar}  & Qwen2.5-VL & 7B & Agent + GRPO  & 42.1 & 56.3 \\
\midrule
\multicolumn{6}{l}{\emph{Ours (Qwen3-VL-4B)}} \\
Qwen3-VL-4B (direct input)         & Qwen3-VL & 4B & VLM                     & 43.5 & --   \\
VLM + RAG Top-3 (no memory)        & Qwen3-VL & 4B & RAG                     & 45.8 & 48.2 \\
AWM-Agent (SFT)                    & Qwen3-VL & 4B & Agent + SFT             & 49.2 & 55.7 \\
\;\;+ Answer-GRPO                  & Qwen3-VL & 4B & GRPO ($R_{\mathrm{ans}}$) & 51.6 & 57.4 \\
\;\;+ AWM-GRPO                     & Qwen3-VL & 4B & GRPO ($R_{\mathrm{AWM}}$) & \textbf{53.9} & \textbf{60.1} \\
\bottomrule
\end{tabular}
\caption{Final-answer accuracy; higher is better. External rows follow their reported protocols and provide context only. Our five Qwen3-VL-4B rows use the official scoring protocol.}
\label{tab:main-results}
\end{table*}

The direct-input and RAG rows provide no-memory baselines, while Sec.~\ref{subsec:analysis} analyzes memory quality only for methods that produce terminal working memory.
On \textsc{MMLongBench-Doc}, AWM-GRPO reaches 53.9 and improves over direct input, RAG Top-3, SFT, and answer-only GRPO by 10.4, 8.1, 4.7, and 2.3 points, respectively.
It also improves over RAG Top-3, SFT, and answer-only GRPO by 11.9, 4.4, and 2.7 points on \textsc{LongDocURL}.

\subsection{Analysis}
\label{subsec:analysis}

\paragraph{Memory-quality analysis.}
Tables~\ref{tab:memory-full} and~\ref{tab:memory-epgiven} report memory-quality metrics across the three AWM-Agent variants on \textsc{MMLongBench-Doc}.
The direct-input and RAG baselines are excluded because they do not produce terminal working memory.
Table~\ref{tab:memory-full} uses the full multi-turn setting; Table~\ref{tab:memory-epgiven} uses the EP-given setting, where retrieval is replaced by gold evidence pages to isolate memory extraction from page access.

% --- Full multi-turn memory-quality table ---
\begin{table}[t]
\centering
\footnotesize
\setlength{\tabcolsep}{4pt}
\begin{tabular}{@{}l c c c@{}}
\toprule
Variant & Final acc. & Mem acc. & $P_{\mathrm{mmc}}$ (\%) \\
\midrule
SFT              & 40.8 & 38.5 & 17.7 \\
+ Answer-GRPO    & 43.2 & 42.5 & 19.9 \\
+ AWM-GRPO       & 45.4 & 44.5 & 17.2 \\
\bottomrule
\end{tabular}
\caption{Memory-quality analysis of the three AWM-Agent variants on 1,082 \textsc{MMLongBench-Doc} trajectories in the full multi-turn setting.}
\label{tab:memory-full}
\end{table}

% --- EP-given memory-quality table ---
\begin{table}[t]
\centering
\small
\setlength{\tabcolsep}{4pt}
\begin{tabular}{@{}l l c c c@{}}
\toprule
Variant & Subset & \shortstack[c]{Final\\acc.} & \shortstack[c]{Mem\\acc.} & \shortstack[c]{$P_{\mathrm{mmc}}$\\(\%)} \\
\midrule
+ Answer-GRPO    & MM EP & 45.4 & 41.2 & 19.1 \\
+ AWM-GRPO       & MM EP & 48.0 & 43.5 & 16.4 \\
\bottomrule
\end{tabular}
\caption{Memory-quality analysis of Answer-GRPO and AWM-GRPO on 733 answerable \textsc{MMLongBench-Doc} examples in the EP-given setting.}
\label{tab:memory-epgiven}
\end{table}

SFT already achieves a relatively low $P_{\mathrm{mmc}}$ of 17.7\%.
Answer-only GRPO improves final-answer accuracy but raises $P_{\mathrm{mmc}}$ to 19.9\%, because it assigns the same reward to memory-supported correct and memory-missing correct trajectories; its effect on memory quality is incidental.
AWM-GRPO separates these two cells through the AWM reward (Sec.~\ref{sec:method}).
Compared with answer-only GRPO, AWM-GRPO improves memory-only accuracy from 42.5 to 44.5 and reduces $P_{\mathrm{mmc}}$ from 19.9\% to 17.2\%, while also improving final-answer accuracy.
AWM-GRPO also improves both final-answer and memory-only accuracy over SFT (+4.6 and +6.0 points) while keeping $P_{\mathrm{mmc}}$ slightly lower (17.2 vs.\ 17.7).
The targeted failure is narrow: after a relevant page is observed, the agent must commit a question-conditioned finding to memory.
AWM-GRPO supplies trajectory-level reward for this terminal-working-memory artifact, rather than changing retrieval tools or the answer prompt.
Table~\ref{tab:memory-epgiven} reports the EP-given comparison: with retrieval replaced by gold evidence pages, AWM-GRPO improves final-answer accuracy from 45.4 to 48.0, memory-only accuracy from 41.2 to 43.5, and reduces $P_{\mathrm{mmc}}$ from 19.1\% to 16.4\% relative to answer-only GRPO, showing that the memory-aware reward remains beneficial even when page access is controlled.

\begin{figure}[t]
\centering
\includegraphics[width=0.85\columnwidth]{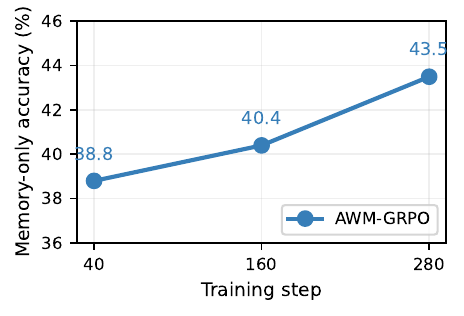}
\caption{AWM-GRPO memory-only accuracy over training steps on \textsc{MMLongBench-Doc}. Memory-only accuracy increases from 38.8 at step 40 to 43.5 at step 280.}
\label{fig:memory-dynamics}
\end{figure}

\paragraph{Training dynamics.}
We track memory-only accuracy across AWM-GRPO training steps on \textsc{MMLongBench-Doc}.
Figure~\ref{fig:memory-dynamics} shows a monotonic increase from 38.8 at step 40 to 43.5 at step 280, suggesting that the reward progressively improves the answerability of terminal working memory during training.

\paragraph{Evidence-source analysis.}
Table~\ref{tab:mmlongbench-source} compares Answer-GRPO and AWM-GRPO by answer-source category.
AWM-GRPO lowers $P_{\mathrm{mmc}}$ for mixed visual+text, text-multi, figure, chart, and none categories, while table and layout questions show higher $P_{\mathrm{mmc}}$.
This suggests that the memory reward helps in several cross-source settings but does not uniformly solve structured-evidence preservation.

\begin{table}[t]
\centering
\small
\setlength{\tabcolsep}{2pt}
\resizebox{\columnwidth}{!}{%
\begin{tabular}{@{}l r cc cc@{}}
\toprule
& & \multicolumn{2}{c}{Final acc.} & \multicolumn{2}{c}{$P_{\mathrm{mmc}}$ (\%)} \\
\cmidrule(lr){3-4} \cmidrule(lr){5-6}
Source & $N$ & Ans & AWM & Ans & AWM \\
\midrule
Pure-text           & 135  & 46.1 & \textbf{58.5} & \textbf{22.8} & 23.1 \\
Table               & 143  & 36.7 & \textbf{39.5} & \textbf{15.2} & 23.9 \\
Chart               & 107  & 45.7 & \textbf{46.0} & 21.4 & \textbf{17.5} \\
Figure              & 187  & 41.0 & \textbf{43.9} & 20.3 & \textbf{16.2} \\
Layout              & 32   & \textbf{75.9} & 74.3 & \textbf{4.8}  & 15.8 \\
visual+text (mixed) & 180  & 36.7 & \textbf{37.3} & 36.2 & \textbf{22.2} \\
visual (multi)      & 3    & --   & --   & --   & --   \\
text (multi)        & 62   & 40.1 & \textbf{48.3} & 31.8 & \textbf{24.0} \\
none                & 233  & 47.5 & \textbf{49.2} & 10.4 & \textbf{5.4}  \\
\midrule
All                 & 1082 & 43.2 & \textbf{45.4} & 19.9 & \textbf{17.2} \\
\bottomrule
\end{tabular}}
\caption{Evidence-source comparison on \textsc{MMLongBench-Doc} in the full multi-turn setting. ``Ans'' denotes Answer-GRPO and ``AWM'' denotes AWM-GRPO; bold marks the better value within each source.}
\label{tab:mmlongbench-source}
\end{table}

\section{Related Work}
\label{sec:related}

\paragraph{Long-document VQA agents.}
\textsc{MMLongBench-Doc}~\citep{mmlongbenchdoc} and \textsc{LongDocURL}~\citep{longdocurl} are two recent benchmarks for long-document VQA over PDFs that average tens to hundreds of pages and mix textual, tabular, and figure-grounded questions.
Doc-V$^{\star}$ treats this task as sequential page navigation: the agent starts from a low-resolution overview, fetches selected page images, and records findings in structured working memory before answering~\citep{docvstar}.
Its navigation policy is trained with imitation learning and GRPO.
MemSearcher studies compact working memory across turns in text-based search agents rather than page-image VQA, using multi-context GRPO to propagate trajectory-level advantages to individual turns~\citep{memsearcher}.
AgenticRAG-R1 uses stack memory for multi-step reasoning, retrieval, and memorizing, while TC-RAG studies Turing-complete RAG for medical LLM systems~\citep{jiang2026agenticrag,jiang2025tc}.
SCAIR, GR-Agent, and EigentSearch structure retrieval or reasoning with schemas or tools~\citep{scair,gragent,eigentsearch,li2025enrich}.
AWM keeps the agent architecture and memory format, but evaluates whether terminal memory is \emph{answerable} on its own and uses that signal as an RL reward.

\paragraph{Intermediate-state supervision.}
TableQA work filters questions and prunes tables to retain answer-critical content~\citep{enotab,tabtrim}.
For visual inputs, CapRL rewards answerable captions, while Vision-SR1 optimizes question-conditioned visual descriptions and answers~\citep{caprl,visionsr1,li2026s,xiao2026staying,xiao2026visual}.
Perception-R1 checks visual content in a reasoning response against reference annotations~\citep{perceptionr1}.
AWM instead evaluates the answerability of query-specific terminal working memory accumulated across sequential page inspections.

\paragraph{Memory evaluation.}
Evidence-citation work asks whether an agent's final answer is supported by its cited evidence~\citep{liu2023evaluating, gao2023enabling}.
Related RAG work tests reasoning when retrieved knowledge is incomplete or exposes the argument structure behind an answer~\citep{whatbreakskgrag,argrag}.
Memory-only answerability asks whether saved memory is sufficient to answer after the page images and interaction trajectory are removed, a question related to self-evaluation in retrieval-augmented generation~\citep{asai2023selfrag}.
Neither test subsumes the other: answerable memory may contain unsupported claims, while grounded memory may omit a fact needed to answer.
Other diagnostics audit prompt-conditioned traces left by RLVR~\citep{diba}; they target data membership, not whether terminal memory supports answering.
In our diagnostic, a frozen language model reads only $(q, M^{\mathrm{term}})$ and produces a memory-only answer, which the official benchmark judge scores alongside the agent's final answer.
This reader-based diagnostic is related to LLM-based evaluation~\citep{zheng2024judge, gilardi2023chatgpt}.
Memory probes in interpretability~\citep{patchscope} ask whether internal activations encode a fact, whereas we ask whether an externalized, source-linked terminal record encodes it.
   % Related Work moved to late position
\section{Conclusion}
\label{sec:conclusion}

Memory-only answerability identifies cases in which an agent answers correctly but leaves terminal working memory that cannot support the answer on its own.
AWM-GRPO improves accuracy over answer-only GRPO on both benchmarks and lowers $P_{\mathrm{mmc}}$ in both settings.
Two benchmarks and one model size leave grounding unverified.
 % Conclusion (was Discussion, folded in F20c)
\clearpage
\section*{Limitations}
\label{sec:limitations}

All controlled AWM comparisons use Qwen3-VL-4B on \textsc{LongDocURL} and \textsc{MMLongBench-Doc}; whether AWM-GRPO helps at 30B+ or on other document distributions, such as scientific papers, financial filings, and slide decks, remains open.
Memory-only answerability may depend on the frozen Qwen3-14B reader, and we did not test robustness with a second reader.
Errors in the fixed judge's answer-extraction step can affect the reported metrics.
Our 100-example audit found six disagreements but no consistent direction of error; a larger audit would provide stronger evidence.
The AWM reward measures answerability, not full source-grounding verification; future work can add source-conditioned checks for individual memory claims.
Memory-only answerability also requires an extra reader pass and an extraction pass during offline evaluation, although neither pass is needed when the trained agent is deployed.

\section*{Ethics Statement}

This work uses publicly released benchmarks (\textsc{MMLongBench-Doc}, \textsc{LongDocURL}), a public open-weight VLM (Qwen3-VL-4B-Instruct), and a fixed local Qwen3-14B model for memory reading and reward scoring during online training.
We collect no new data, do not work with human subjects, and do not target deployment.
The official judge follows each benchmark's answer-extraction and rule-based scoring pipeline.
Separately, the controlled intervention in Appendix~\ref{app:oracle} uses GPT-4o to construct improved memory from gold evidence pages; those memories are not used for training.
The fixed training scorer, reader, and answer extractor may introduce systematic errors; our manual audit does not establish broader scoring robustness or claim-level source grounding.
Per ARR / EMNLP policy, AI assistants were used for editing; all technical content is the authors' own.

% acl.sty already sets \bibliographystyle{acl_natbib} internally; do not re-declare here.
\bibliography{bibliography}

\appendix
\section{Dataset Details}
\label{app:datasets}

\paragraph{Benchmark sources.}
\textsc{MMLongBench-Doc} uses all 1,082 examples in the original \texttt{samples.json}. \textsc{LongDocURL} uses all 2,325 examples in the original \texttt{LongDocURL\_public.jsonl}.
The main-results table uses the original full-size \textsc{MMLongBench-Doc} page images.

\paragraph{EP-given evaluation.}
In the diagnostic experiments, EP-given replaces retrieval with the annotated gold evidence pages and is reported on fixed \textsc{MMLongBench-Doc} subsets specified in the corresponding tables.
We do not report a \textsc{LongDocURL} EP-given diagnostic in the main experiments.

\paragraph{Routing verification.}
All reported runs pass a check that each question is paired with its source document.

\section{Evaluation Protocol}
\label{app:judge}

\paragraph{Official benchmark scoring.}
All final-answer and memory-only metrics reported for our Qwen3-VL-4B runs on \textsc{MMLongBench-Doc} and \textsc{LongDocURL} use the official judge $J$ defined in Sec.~\ref{sec:diagnostic}.
The judge follows the official evaluation pipeline for the corresponding benchmark.
It first sends the model's free-form output to GPT-4o at temperature $0$.
GPT-4o extracts a concise candidate answer in the benchmark's expected format.
A benchmark-specific deterministic rule scorer then compares the extracted candidate with the gold answer and assigns the official generalized-accuracy score.

\paragraph{Type-aware rule scoring.}
Both official scorers normalize the extracted candidate and gold answer to the benchmark's answer types before comparison.
Integer answers require an exact match.
Floating-point answers allow a $1\%$ relative tolerance and account for equivalent percentage scales.
For string and \texttt{None} answers, the scorer uses exact matching for designated special strings and otherwise uses average normalized Levenshtein similarity (ANLS) with a $0.5$ threshold.
For list answers, the \textsc{MMLongBench-Doc} scorer requires equal list lengths, sorts both lists, and uses the minimum element-level ANLS as the question score.
The \textsc{LongDocURL} scorer instead matches list elements greedily by ANLS and applies a square-root penalty when the list lengths differ.

\paragraph{Final-answer and memory-only metrics.}
For final-answer accuracy, the candidate is the agent's final answer.
For memory-only accuracy, a frozen reader receives the question $q$ and terminal working memory $M^{\mathrm{term}}$, but no trajectory, page images, or tool instructions.
It then produces the candidate answer.
The reader is the frozen Qwen3-14B model defined in Sec.~\ref{sec:diagnostic}.
The same judge evaluates both candidates.
Unless stated otherwise, reported accuracy is the mean official per-question generalized-accuracy score.
The four-cell memory analysis requires binary outcomes, so it treats any positive official score as correct and a zero score as incorrect.
The controlled intervention in Sec.~\ref{sec:oracle-intervention} reports this binary positive-score rate so that each percentage matches its correct count and Wilson confidence interval.

\paragraph{Manual audit.}
We manually checked 100 scored examples against human correctness judgments.
The audit found six disagreements, mainly involving answer aliases, list formatting, and partial-credit edge cases.
The six disagreements were not concentrated on one side of the comparison.

\paragraph{Training-time reward.}
Online GRPO uses a frozen local Qwen3-14B model to generate the memory-only answer and compute the two reward scores.
This training-time scorer is separate from the official judge used for reported metrics.
Only the Qwen3-VL-4B agent policy is updated; the Qwen3-14B model remains fixed.

\section{Design Choices and Reward Variants}
\label{app:design}

This appendix expands on the design choices and reward analyses summarized in the main text.

\subsection{Why a conditional rate and not a marginal gap}

For a set of $N$ examples, let $n_{10}$ and $n_{01}$ denote the numbers of memory-missing-correct and answering-error examples, respectively.
The marginal gap between final-answer and memory-only accuracy is $(n_{10}-n_{01})/N$.
Because the two failures enter with opposite signs, they can cancel: the gap can be near zero even when many correct answers have unanswerable terminal working memory.
It therefore does not directly answer the question that motivates AWM: among the final answers that the agent gets right, how often is the saved memory still insufficient?

The conditional rate $P_{\mathrm{mmc}}=n_{10}/N_{\mathrm{correct}}$ measures the fraction of binary-correct final answers with unanswerable terminal working memory.
Answering-error examples enter neither its numerator nor its denominator, so they cannot cancel the memory failures of interest.
Across datasets, subsets, and checkpoints, $P_{\mathrm{mmc}}$ retains the same interpretation: the share of correct final answers that cannot be recovered from memory alone.
We therefore report it rather than the signed marginal gap.

\subsection{Why mandatory commit does not collapse memory-only accuracy onto final-answer accuracy}

Our agent loop makes a memory update mandatory after every retrieval.
A natural worry is that this construction makes memory-only accuracy track final-answer accuracy because every retrieval is followed by a memory update.
It does not.
Each update is free-form within a schema that limits findings to 40 words and requires a source page.
An agent can satisfy that schema with a generic page description, omit the fact needed by the question, and still answer from raw page context at the final step.
The diagnostic in Sec.~\ref{sec:diagnostic} tests this gap between writing an update and preserving what the later answer needs.
Mandatory commit therefore isolates \emph{what} the agent extracts at each update rather than \emph{when} it chooses to update memory.

\subsection{Reward variants we considered}

The minimal form in Sec.~\ref{sec:method} pairs one answer from the frozen reader with one score for that answer.
We also considered three richer variants.

\paragraph{Per-finding source-image grounding.}
A more granular signal asks a model to check, for each finding in $M^{\mathrm{term}}$, whether the cited \texttt{source\_page} image actually supports the claim.
This is a stricter per-finding evidence-grounding signal and would provide a finer-grained training signal.
We did not adopt it for the main experiments because (i) it requires one scoring call per finding rather than one per trajectory, raising compute and scoring noise at training time, and (ii) it conditions the reward on a fixed schema, sacrificing the schema-agnosticism that we view as a feature of our minimal form.
Source-grounding audits using this variant are a natural follow-up evaluation, separate from the reward signal used during training.

\paragraph{Self-consistency over multiple reader rollouts.}
The binary memory score inherits the variance of a single reader rollout.
Sampling $K$ answers from $R(q, M^{\mathrm{term}})$ and scoring their correct fraction would reduce variance but multiply scoring cost by $K$; we do not evaluate this variant.

\paragraph{Length penalty on $M^{\mathrm{term}}$.}
We considered a training-time penalty for $M^{\mathrm{term}}$ entries that closely mirror the predicted final-answer string.
We did not adopt it because such a penalty could suppress legitimate answer-supporting findings that contain the answer value (e.g., when the answer is a value on the page); answer-echo detection remains a separate audit.

\subsection{Computation overhead relative to standard GRPO}

\paragraph{Training-time overhead.}
Relative to standard answer-only GRPO, AWM-GRPO adds two short text-only passes through a local Qwen3-14B model for each rollout.
The first pass answers from terminal working memory alone, and the second scores that memory-only answer for the training reward.
The answer-only baseline already scores the agent's final answer.
The Qwen3-14B model remains frozen, and only the Qwen3-VL-4B agent policy is trained.
We report the added model passes instead of a wall-clock percentage because elapsed time depends on hardware, batching, and serving configuration.

\paragraph{Deployment-time overhead.}
The memory reader and training scorer are not called at deployment.
AWM-GRPO therefore adds no deployment-time model call relative to the same agent policy and tool loop.

\subsection{Reward discrimination under AWM}
\label{app:reward-discrimination}

This appendix gives a structural statement of how the AWM reward (Sec.~\ref{sec:method}) differs from an answer-only reward over the four outcome cases of Table~\ref{tab:cells2x2}.
The statement is purely structural and assumes only that the scoring and reader interfaces are fixed within each pass.
During online training, the frozen local Qwen3-14B model supplies the reader output and the two reward scores.
For reported diagnostics, $R$ is the frozen Qwen3-14B reader and $J$ is the official judge from Sec.~\ref{sec:diagnostic}.

Let $s_{\mathrm{ans}}, s_{\mathrm{mem}} \in \{0,1\}$ denote final-answer correctness and memory-only answerability for a trajectory $\tau$, as in Sec.~\ref{subsec:reward}, and let $R_{\mathrm{AWM}}$ be the piecewise AWM reward with values $(r_{11}, r_{10}, r_{01}, r_{00})$ satisfying $r_{11} > r_{10} > r_{01} > r_{00}$.
The reward mapping $(s_{\mathrm{ans}}, s_{\mathrm{mem}}) \mapsto R_{\mathrm{AWM}}$ strictly orders the four outcome cells: memory-supported correct receives the highest value, memory-missing correct sits above answering error, and unresolved error receives the lowest value.
Answer-only reward, by contrast, takes only two values and identifies $(s_{\mathrm{ans}}, s_{\mathrm{mem}}){=}(1,1)$ with $(s_{\mathrm{ans}}, s_{\mathrm{mem}}){=}(1,0)$ and identifies $(0,1)$ with $(0,0)$.
The AWM reward therefore distinguishes pairs of policies that agree on final-answer accuracy but differ in memory-only answerability; answer-only reward cannot.
The default values $(2,\ 0,\ -0.1,\ -1)$ also preserve final-answer priority because $r_{10} > r_{01}$, so a final-answer-correct trajectory with unanswerable memory is never outranked by a final-answer-wrong trajectory with answerable memory.

The same property has a direct GRPO consequence~\citep{xie2026unlocking,xie2026edgeexperiencedistillationguidedexploration,zhou2026look,zhou2026one}.
In a GRPO group of $G$ trajectories sampled for one prompt, sampled trajectories that fall into different reward levels can produce non-zero group-relative advantage after normalization.
The key case is a group containing both memory-supported correct and memory-missing correct trajectories: answer-only reward assigns them the same value, while the AWM reward assigns values that differ by $r_{11} - r_{10}$.
By the diagnostic of Sec.~\ref{sec:diagnostic}, this is the cell pair where the pre-training agent already shows a measurable memory-preservation gap.

\subsection{Advantage-distribution simulation details}
\label{app:advantage-sim}

This appendix records the simulation behind Figure~\ref{fig:advantage-dist}.
For mixture $m$, let $\rho^{(m)}$ be a categorical distribution over the four cells $z=(s_{\mathrm{ans}},s_{\mathrm{mem}})$.
For each reward scheme and mixture, we sample $B=15{,}000$ groups of size $G=8$.
AWM samples $z_{b,1:G}\sim \mathrm{Categorical}(\rho^{(m)})$, while answer-only sampling uses the corresponding final-answer marginal.
We assign either answer-only or AWM rewards, compute the group mean $\mu_b$ and standard deviation $\sigma_b$, and form $A_{b,i}=(r_{b,i}-\mu_b)/(\sigma_b+\epsilon)$ with $\epsilon=10^{-6}$.
Figure~\ref{fig:advantage-dist} shows kernel-smoothed empirical advantage distributions grouped by final-answer correctness for answer-only reward and by outcome cell for AWM.
Finite groups and discrete rewards produce a finite set of normalized advantage values; kernel smoothing affects only the plot, not reward or training computation.
In $(0,0),(0,1),(1,0),(1,1)$ order, the three mixtures are $(0.82,0.05,0.05,0.08)$, $(0.60,0.05,0.05,0.30)$, and $(0.30,0.05,0.05,0.60)$.
The off-diagonal cells $(0,1)$ and $(1,0)$ therefore retain probability $0.05$ across rows, while diagonal mass shifts from $(0,0)$ toward $(1,1)$.
The change in $(1,0)$ advantage therefore comes from the rising group mean rather than a change in its cell frequency.

\section{Improved-Memory Intervention Details}
\label{app:oracle}

This appendix expands the controlled memory comparison in Sec.~\ref{sec:oracle-intervention}.
The intervention is a controlled sanity check rather than a deployed method or a full-benchmark result, and the improved memories are not used as training data.
The intervention uses a fixed set of 500 answerable \textsc{MMLongBench-Doc} questions in the evidence-page-given (EP-given) setting.
Here, \emph{answerable} means that the benchmark annotations identify sufficient evidence for a reference answer.
It does not mean that the agent's terminal working memory preserves that evidence.

\subsection{Setup}

EP-given evaluation provides the gold evidence pages, which removes page retrieval as a source of error.
The empty condition supplies no terminal memory.
The original condition uses memory written by the base Qwen3-VL-4B agent, while the improved condition uses memory that GPT-4o constructs from the same gold evidence pages.
The same 500 questions are evaluated under all three terminal-memory conditions.
The final-answer generator and judge are held fixed across conditions.
For memory-only evaluation, the fixed reader is Qwen3-14B.

All reported outputs use the official judge defined in Sec.~\ref{sec:diagnostic}.

The original agent partitions the 500 examples into the four outcome cells from Sec.~\ref{sec:diagnostic}: 104 memory-supported correct, 77 memory-missing correct, 47 answering error, and 272 unresolved error.
We fix this partition from the original-memory run, then compare the same questions under original and improved memory.

\subsection{Final-answer binary correctness}

\begin{table}[t]
\centering
\small
\resizebox{\columnwidth}{!}{%
\begin{tabular}{l c c c}
\toprule
Memory condition & Correct / N & Rate (\%) & 95\% Wilson CI (\%) \\
\midrule
Empty memory               & 14 / 500  & 2.8  & [1.7, 4.6] \\
Original agent memory (4B) & 181 / 500 & 36.2 & [32.1, 40.5] \\
Improved memory (GPT-4o)   & 222 / 500 & 44.4 & [40.1, 48.8] \\
\bottomrule
\end{tabular}}
\caption{Final-answer positive-score rate in the controlled memory intervention. Intervals are 95\% Wilson confidence intervals.}
\label{tab:oracle-main}
\end{table}

Under the shared evaluation procedure, final-answer binary correctness is 44.4\% with improved memory and 36.2\% with original agent memory.
With empty memory, 14 of 500 final answers receive a positive official score.
These rates describe the controlled 500-example comparison and are not a full-benchmark accuracy claim.

\subsection{Memory-only results by original outcome}

\begin{table}[t]
\centering
\small
\resizebox{\columnwidth}{!}{%
\begin{tabular}{l c c c c}
\toprule
Original outcome & N & \shortstack{Original memory\\correct} & \shortstack{Improved memory\\correct} & \shortstack{Change\\(count; pp)} \\
\midrule
Memory-supported correct & 104 & 104 & 88 & $-16$; $-15.4$ \\
Memory-missing correct   & 77  & 0   & 48 & $+48$; $+62.3$ \\
Answering error          & 47  & 47  & 29 & $-18$; $-38.3$ \\
Unresolved error         & 272 & 0   & 59 & $+59$; $+21.7$ \\
\midrule
Total                    & 500 & 151 & 224 & $+73$; $+14.6$ \\
\bottomrule
\end{tabular}}
\caption{Memory-only correctness after grouping examples by the original agent's outcome. Change is measured within each fixed row.}
\label{tab:oracle-cells}
\end{table}

The original-memory column follows directly from the fixed outcome labels: memory is answerable in the memory-supported-correct and answering-error cells, and unanswerable in the other two cells.
Improved memory makes 48 of the 77 memory-missing-correct examples and 59 of the 272 unresolved-error examples answerable.
The same replacement makes memory unanswerable for 16 originally memory-supported-correct examples and 18 answering-error examples.
The net result is 73 more memory-answerable examples, an increase from 151 to 224 out of 500.
The intervention therefore improves memory-only answerability overall but does not dominate the original memory in every outcome cell.

\subsection{Artifact audit}

All 500 improved memories include a \texttt{source\_page} citation to a page in the corresponding gold \texttt{evidence\_pages} set.
This audit verifies that every memory points to an eligible evidence page.
It does not verify that every generated statement is entailed by that page or that the memory contains all evidence needed for the answer.

We also apply a strict answer-string filter.
The filter removes every example whose improved memory contains the normalized gold answer string after whitespace and punctuation normalization.
It leaves 304 examples, of which the memory-only reader receives a positive official score on 98, for a positive-score rate of 98 / 304 = 32.2\%.
This remaining rate shows that exact answer-string copies do not explain all successful memory-only answers.
Because the filter selects a different subset rather than editing memory in place, its 32.2\% rate is not a like-for-like estimate of the full-set effect.
The filter also cannot rule out paraphrased answer leakage or establish claim-level grounding.

\subsection{Answerability versus grounding}

Memory-only answerability asks whether the fixed reader can recover an accepted answer from the question and terminal working memory alone.
Grounding asks whether the memory's claims are supported by the cited source pages.
These properties are related but not equivalent.
An answerable memory may contain an unsupported guess or an answer echo.
A grounded memory may still omit a needed fact, or the reader may fail to use it.
The audits check only source-page validity and exact answer copying; they do not establish claim-level grounding.
Accordingly, the intervention shows that changing memory content can improve answering under a fixed protocol; it does not show that every improved memory is fully grounded.

\subsection{Limitations}

This single-seed intervention is an upper-bound analysis on one fixed subset, reader, judge, and improved-memory model: GPT-4o is stronger than the original 4B memory writer and receives gold evidence pages.
Table~\ref{tab:oracle-cells} shows that improved memory also hurts some outcome cells, and the improved memories are used neither for training nor in the main AWM diagnostic.
Repeated runs and claim-level audits are needed.

\section{Relation to Other Reliability Criteria}
\label{app:reliability}

AWM treats terminal working memory as an instance-level artifact.
Predictive multiplicity in knowledge-graph embeddings and disagreement across repeated language-model queries expose cases where aggregate performance hides query-level variation~\citep{zhu-etal-2024-predictive,RobustLLM}; AWM instead tests whether a fixed reader can recover the answer after the pages and interaction trace are removed.
Conformal methods instead construct prediction sets or study coverage under predicate and configuration shift~\citep{zhu-etal-2025-conformalized,zhu-etal-2025-predicate,zhu2026conformalized}.
We make no uncertainty or coverage claim; the output is a per-trajectory memory-only answerability test.

Other studies modify or inspect the evidence-to-reasoning path.
EigentSearch-Q+ uses structured tools and SCAIR uses schema-conditioned traversal~\citep{eigentsearch,scair}.
Knowledge-graph RAG studies test missing facts, while ArgRAG represents retrieved evidence as supporting or attacking arguments~\citep{zhou2025evaluating,whatbreakskgrag,argrag}.
Related analysis also finds asymmetric responses for logically equivalent facts under different pretraining frequencies~\citep{he2025supposedly}.
AWM leaves the agent architecture and memory schema unchanged and evaluates the source-linked memory they produce: the memory passes only when the reader can answer without the pages or the trajectory.

\end{document}